%% file: main.tex
\documentclass[preprint,12pt,nopreprintline]{elsarticle}

\usepackage{amssymb}
\usepackage{amsmath}

\usepackage{bm}
\usepackage{url}
\usepackage{tabularx}
\usepackage{xcolor}
\usepackage{soul}
\usepackage{xcolor} 
\sethlcolor{yellow} 

\usepackage[normalem]{ulem}

\journal{}

\begin{document}

\begin{frontmatter}



\title{The Impact of Item-Writing Flaws on Difficulty and Discrimination in Item Response Theory} 

\author[label1]{Robin Schmucker}
\author[label2]{Steven Moore}

\affiliation[label1]{organization={Machine Learning Department},
             addressline={Carnegie Mellon University},
             city={Pittsburgh},
             state={PA},
             country={USA}, email={, rschmuck@cs.cmu.edu}}

 \affiliation[label2]{organization={Human-Computer Interaction},
             addressline={Carnegie Mellon University},
             city={Pittsburgh},
             state={PA},
             country={USA},
             email={, StevenJamesMoore@gmail.com}}

%


\begin{abstract}
High-quality test items are essential for educational assessments, particularly within Item Response Theory (IRT). Traditional validation methods rely on resource-intensive pilot testing to estimate item difficulty and discrimination. More recently, Item-Writing Flaw (IWF) rubrics emerged as a domain-general approach for evaluating test items based on textual features. This method offers a scalable, pre-deployment evaluation without requiring student data, but its predictive validity concerning empirical IRT parameters is underexplored. To address this gap, we conducted a study involving 7,126 multiple-choice questions across various STEM subjects (physical science, mathematics, and life/earth sciences). Using an automated approach, we annotated each question with a 19-criteria IWF rubric and studied relationships to data-driven IRT parameters. Our analysis revealed statistically significant links between the number of IWFs and IRT difficulty and discrimination parameters, particularly in life/earth and physical science domains. We further observed how specific IWF criteria can impact item quality more and less severely (e.g., negative wording vs. implausible distractors) and how they might make a question more or less challenging. Overall, our findings establish automated IWF analysis as a valuable supplement to traditional validation, providing an efficient method for initial item screening, particularly for flagging low-difficulty MCQs. Our findings show the need for further research on domain-general evaluation rubrics and algorithms that understand domain-specific content for robust item validation.
\end{abstract}




\end{frontmatter}



\section{Introduction}
\label{sec:introduction}

Multiple-choice questions (MCQs) are recognized as an effective and commonly used form of assessment across diverse educational domains. However, developing high-quality MCQs remains a time-intensive process that often relies heavily on expert judgment and post-deployment student data analysis~\citep{Rashwan2024:Postexamination}. This creates significant bottlenecks in test development workflows, particularly for large-scale assessments where hundreds or thousands of items must be evaluated efficiently. Ensuring these questions are of high quality is critical for maintaining validity, reliability, and overall soundness of assessing student learning \citep{mccoubrie2004improving,boland2010writing}. In both standardized testing (e.g., GRE, MCAT, SAT) and classroom assessments, rigorous evaluation is applied to retain only the most reliable MCQs \citep{elgadal2021item}. This process allows educators and researchers to make targeted improvements, revising or discarding flawed items to better measure student learning. Among the established methods for evaluating MCQ quality, Item Response Theory (IRT) is often considered the gold standard \citep{mulla2023automatic}. By quantifying item performance through parameters such as difficulty and discrimination, IRT provides valuable insights into how students interact with different questions.

While IRT has proven effective at capturing statistical dimensions of item performance, it does not fully explain \textit{why} certain questions might vary in difficulty or discrimination. It requires substantial student response data and operates post hoc, often identifying poor-quality questions only after students have encountered them \citep{rusch2017breaking}. Additionally, IRT parameters may overlook qualitative aspects of question design, such as pedagogical soundness and specific design flaws that decrease assessment integrity~\citep{Haladyna2004:Construct}. Expert review and rubric-based evaluations help address these limitations by detecting specific question design flaws that may skew assessment outcomes \citep{kurdi2020systematic,bates2014assessing}. While researchers acknowledge that such flaws influence item performance, a systematic examination of how qualitative shortcomings in item design interact with quantitative IRT measures across different domains remains limited. Empirical evidence linking specific flaws to changes in item discrimination and difficulty could not only clarify why certain questions perform poorly, but also enable the development of automated screening tools that flag potentially problematic items before they reach students. This would present a significant advancement in assessment efficiency, allowing educators and test developers to identify and revise flawed and potentially harmful questions during the design phase rather than after costly data collection.

To address this gap, the present study integrates IRT analysis with the standardized Item-Writing Flaws (IWF) rubric \citep{tarrant2006frequency}, an instrument for expert evaluation of MCQ quality. To explore relationships between IRT- and IWF-based evaluations, we analyze datasets across diverse educational domains: life and earth sciences, physical sciences, and mathematics, encompassing the middle and high school grade levels in the United States. These datasets combine 7,126 MCQs with response data of 448,000 students within a large-scale online learning platform. For each question, we assess difficulty and discrimination parameters and automatically apply the 19-criterion IWF rubric. By comparing these quantitative and qualitative evaluations, we aim to demonstrate how specific design flaws influence item performance across subject domains. We investigate three primary research questions:
\begin{description}
\item[RQ1] How does IWF frequency correlate with IRT difficulty and discrimination parameters across MCQs from different educational domains?
\item[RQ2] Which IWF criteria are most strongly associated with low-quality questions, as indicated by IRT difficulty and discrimination parameters?
\item[RQ3] To what extent can IWF-based item features be leveraged to (i) predict IRT difficulty and discrimination parameters and (ii) filter out low-quality items?
\end{description}

These research questions directly address the practical needs of assessment developers who must balance quality assurance with efficiency constraints, providing a pathway toward more systematic and scalable item evaluation processes. Beyond generating insights into the connections between IRT and IWFs, this research introduces a scalable methodology that can transform current assessment development practices. By demonstrating how AI-enabled qualitative evaluations can be systematically integrated with statistical measurements from student data, we provide assessment designers with a practical framework for early identification of problematic items. This approach offers the potential to significantly reduce development costs, improve assessment quality, and accelerate the creation of more effective and equitable educational measurements. The resulting hybrid framework equips educators, test developers, and researchers with actionable, evidence-based strategies to enhance their assessment practices while maintaining the rigor expected in educational measurement.

\section{Related Work}
\label{sec:related_work}

Here, we survey key developments and methodologies in automated evaluation, IRT-based item validation, and rubric-based learning science evaluations, providing the context upon which the present study builds.

\subsection{Automated Evaluation Techniques}

Automated evaluation of MCQs is increasingly essential for large-scale assessment, both to lighten the burden on human reviewers and to screen items before they reach students~\citep{gorgun2024exploring}. Traditional automatic metrics such as BLEU and ROUGE often fail to capture the nuances of question quality, so recent work has turned to more comprehensive linguistic and contextual features~\citep{al2024analysis}. The BEA’24 shared task on automated prediction of MCQ difficulty and response time provided the first community benchmark and showed that transformer ensembles can outperform traditional baselines when modeling both student error rates and latency~\citep{yaneva2024findings}. Classic psychometric measures, such as IRT difficulty and discrimination, continue to serve as foundational metrics for question quality assessment~\citep{rezigalla2024item}. However, these methods require post-hoc student data, so researchers often supplement them with other automated quality measures such as diversity, answerability, and cognitive complexity~\citep{moore2024automatic}. For example, \citet{scaria2024automated} prompted LLMs to generate items at specific levels of Bloom's taxonomy~\citep{Bloom1956:Handbook} and found that expert raters judge higher-order questions to be as valid as baseline recall items, even though automated scores still trail human judgment. Other efforts have proposed zero-shot methods for using LLMs to evaluate MCQ quality~\citep{al2024analysis}. While these approaches show promising results, their performance remains inconsistent, with human evaluation often serving as the gold-standard comparison. To narrow this gap, the MIRROR framework employs an iterative committee of LLM agents to rate relevance, novelty, complexity and grammar, increasing GPT-to-expert correlations to 0.83 without any domain experts in the loop~\citep{deroy2024mirror}. While this approach approaches human-level evaluation, it is limited to assessing grammar, appropriateness, complexity, and novelty. A significant gap in current evaluation methods, particularly those using LLMs, is their narrow focus on subjective quality over essential pedagogical considerations and cognitive load.

Beyond expert and LLM review, IRT uses student data to determine question difficulty and discrimination, which are parameters central to assessment integrity~\citep{Ayala2013:Theory}. As an alternative to post-hoc analysis of student data, recent NLP methods aim to predict IRT parameters directly from the question text. One study enriches each MCQ with chain-of-thought rationales and synthetic ability profiles, cutting mean-squared error in difficulty prediction by 28\% and aligning option-level likelihoods with a 2-PL IRT target~\citep{feng2025reasoning}. AnaQuest is a prompting technique that injects students’ expressed misconceptions into GPT-4, producing foils whose empirical difficulty and discrimination mirror human-written items when analysed with IRT on classroom data~\citep{shimmei2025tell}. Additionally, researchers have begun evaluating psychometric properties of AI-generated assessment items, establishing quality benchmarks for automated question creation~\citep{bhandari2024evaluating}. Longitudinal studies of MOOC physics courses further show that automatically calibrated parameters stay stable across course re-runs and highlight items whose wording changes erode validity, underscoring the practical value of end-to-end IRT pipelines for large-scale online learning environments~\citep{gershon2024evaluation}. This is critical, as a related study has even identified clear and glaring flaws in MCQs used in production courses on popular MOOC platforms~\citep{costello2018future}.

\subsection{IRT-based Item Validation}

Item validation is critical to ensure that assessments measure intended constructs, have appropriate difficulty and discrimination properties, and provide fair evaluations across diverse populations, such as gender and age groups~\citep{Haladyna2013:Developing}. Within the IRT framework, item validation involves conducting pilot studies to collect sufficient student response data for reliable parameter estimation--an expensive and time-consuming process \citep{Baker2001:Basics}. Additionally, if the questions are flawed and have improper difficulty levels, they can hinder and misrepresent student learning. To reduce the amount of student data needed for reliable estimation, recent work proposed multi-armed bandit algorithms as a more data-efficient approach to adaptively refine item parameters~\citep{Sharpnack2024:Banditcat}. Additionally, to warrant fairness and equity of assessments, differential item functioning (DIF) has been used to ensure that items do not advantage or disadvantage any particular group of test-takers~\citep{Ayala2013:Theory}.

As an alternative to student data-driven validation methods, researchers have explored natural language processing (NLP) techniques to predict IRT parameters based on an item's syntactic and semantic features~\citep{Alkhuzaey2024:Text}. Several studies have applied neural networks, such as LSTMs and transformers, to analyze item text and estimate discrimination and difficulty~\cite{Byrd2022:Predicting, Benedetto2023:Quantitative, Reyes2023:Multiple}. These predictions can help mitigate the cold-start problem, reducing the amount of student response data needed for reliable parameter identification~\citep{Mccarthy2021:Jump}. In parallel with the development of the Duolingo English Test, researchers have introduced methods to accelerate IRT parameter initialization, iterative calibration, and assessment item validation~\citep{Yancey2024:Bert, sharpnack2024autoirt, Sharpnack2024:Banditcat}.

The present study utilizes large-scale data of students interacting with 7,126 MCQs and employs an automated approach that combines rule-based methods and last-step LLM verification to annotate each question with a 19-criteria Item-Writing Flaw (IWF) rubric~\citep{tarrant2006frequency}. Unlike prior work focused on improving the \textit{accuracy} of IRT parameter prediction from textual features, our study aims to enhance our \textit{understanding} of relationships between IWF- and IRT-based validation methods, providing actionable insights into how linguistic characteristics influence psychometric properties.

\subsection{Learning Science Rubrics}

Rubrics play a central role in education by providing a structured means of evaluating quality, whether in student submissions or instructional and assessment materials~\citep{allen2006rubrics}. When applied by trained individuals, rubrics help ensure consistency and replicability by offering standardized and interpretable evaluation criteria~\citep{kind2019development}. As a result, rubrics have been used to assess the quality of student-facing resources, including hints, short-answer questions, and multiple-choice questions (MCQs) \citep{price2019comparison, horbach2020linguistic, mulla2023automatic}. For example, \citet{arif2024generation} employed six question-level metrics, including relevance, answerability, and difficulty, to evaluate the quality of LLM-generated MCQs. However, some rubric criteria involve a degree of subjectivity, such as \textit{relevance}, which may affect inter-rater reliability and make replication more challenging. Factors such as language preference, prior knowledge, and personal definitions of difficulty may lead to inconsistencies in applying the same criteria~\citep{smith2016use}. Additionally, even relatively short rubrics can be time-consuming and cumbersome to apply across large content pools, limiting scalability~\citep{haladyna2002review}.

Despite the challenges of rubric-based evaluations, one prominent instrument that has been widely adopted for MCQ assessment is the IWF rubric~\citep{haladyna2002review, tarrant2006frequency, costello2018future}. Applicable across subject areas, the IWF rubric detects flaws such as gratuitous detail, grammatical cues, and implausible or disproportionately long distractors. Two previous studies in the domain of medical education have shown that the presence of IWFs correlates with psychometric properties such as difficulty and discrimination, with flawed items introducing construct-irrelevant variance that can disadvantage students and reduce test validity~\citep{downing2005effects, rush2016impact}. Compared to simpler automated measures (e.g., diversity, perplexity), the IWF rubric provides a more targeted and pedagogically grounded assessment of MCQ quality~\citep{Moore2024:Automatic}. To address the time-intensive nature of manually applying the IWF’s 19 distinct criteria to each MCQ, recent research has focused on automating the process, enabling IWF rubric application at scale~\citep{moore2023assessing, Moore2024:Automatic}. The automated approach achieved an overall accuracy of 94\% on a dataset of 271 MCQs spanning five educational domains, each annotated with a gold-standard expert evaluation of the 19 IWFs. In addition to accelerating the evaluation process, this method enhances assessment consistency by mitigating some of the inherent subjectivity in human-applied rubrics, as many of the criteria are purely objective~\citep{peeters2015measuring}.

While rubrics are widely used in education, they often lack quantitative evidence to demonstrate their effectiveness~\citep{janssen2015building}. In this work, we address this gap by providing quantitative proof that the IWF rubric criteria influence both question difficulty and discrimination. Unlike previous research that compared automatically identified IWF criteria with human-applied labels, our approach relies on an automated application that has already been validated~\citep{moore2023assessing, Moore2024:Automatic}. Consequently, we apply these verified annotation methods to thousands of real MCQs drawn from a variety of domains, using student response data to go beyond mere frequency counts. This enables us to offer actionable insights into how specific IWFs differentially affect question quality.

\section{Methodology}
\label{sec:methodology}

\subsection{Study Context and Dataset}
\label{subsec:dataset}

Our analyses utilize a dataset from the CK-12 Foundation, a US-based nonprofit that provides millions of students with free access to educational resources. CK-12 actively develops and hosts the FlexBook 2.0 system\footnote{\url{https://www.ck12.org}}, an online tutoring platform offering courses across diverse subjects and grade levels. Each course consists of a series of concepts, analogous to a textbook chapter and typically consisting of one to four learning objectives. Each concept is associated with a broader lesson topic and with a practice section designed to develop and assess students’ understanding of that concept. For instance, in a Life Science course, a lesson might be ``Genetics,'' with ``Punnett Squares'' as a concept within it. We focus on popular concepts within middle and high school courses, spanning topics in physical sciences (physics, chemistry), mathematics (algebra, geometry), and life and earth sciences, using data collected from 2023 and 2024.

\input{Table_1.tex}

Overall, our study uses data from 448,000 students interacting with 13,158 questions to determine IRT parameters for 1,033 distinct concepts (Table~\ref{table_1}). All questions were written by human experts. As the Item-Writing Flaw (IWF) rubric studied in this paper is designed specifically for multiple-choice questions (MCQs)~\citep{tarrant2006frequency}, we assess the relationships between IWFs and question-specific difficulty and discrimination parameters based on the 7,126 MCQs within the content pool. The following discusses the IRT parameter estimation and IWF annotation processes in detail.

\subsection{Item Response Theory}
\label{subsec:irt_estimation}

Item Response Theory (IRT) is a methodological framework commonly used in high-stakes assessments, such as college entrance exams, such as the SAT and GRE~\citep{Ayala2013:Theory}. Formally, IRT models interactions between students and a set of test items (questions) under binary ($\{0, 1\}$) and polytomous (partial credit scoring, Likert scale, etc.) response outcomes. The idea is to assign each student a latent ability parameter that explains their observed response outcomes, estimated using probabilistic inference. The relationship between student ability and response outcome probabilities is modeled by fitting appropriate item response functions (IRFs) for each item--sigmoid functions for binary items, or more complex functions like the partial credit model or graded response model for polytomous items~\citep{Ayala2013:Theory}. In this paper, we focus on binary outcomes and sigmoid functions, as all MCQ responses are graded as either ``correct'' or ``incorrect''.

For each item $i$, its IRF reaches its steepest slope at a specific point on the $x$-axis, representing the item's difficulty $\delta_i$. The steepness of the IRF reflects the item’s discrimination property, denoted as $\alpha_i$. Given student ability $\theta_s$, along with item difficulty $\delta_i$ and discrimination parameters $\alpha_i$, the probability of student $s$ answering item $i$ correctly is defined as
\begin{align}
\mathbb{P}(X_{si}=1 \,|\, \theta_s,\, \alpha_i,\, \delta_i) = \frac{1}{1-e^{-\alpha_i(\theta_s - \delta_i)}},
\label{eq_irt}
\end{align}
where $X_{si}$ indicates the binary response outcome. The item response matrix $X$ captures all interactions between students and items. In real-world domains, $X$ can contain missing entries because not all students respond to all items. However, the IRT estimation process naturally accommodates this sparsity when estimating model parameters using maximum likelihood estimation based only on the observed responses.

Our study utilizes the R package MIRT~\citep{Chalmers2012:Mirt} to estimate a separate set of IRT parameters for each of the 1,033 concepts within our dataset. Following guidance from~\citet{Ayala2013:Theory} and~\citet{Baucks2024:Gaining}, we ensure robust estimation of IRT parameters by focusing on concepts that meet specific response thresholds: each concept must have data from at least 500 students, with each student providing responses to a minimum of 5 questions, and each question must have responses from at least 500 students. Adhering to these criteria mitigates potential sparsity issues in our item response data, thereby promoting the reliability and accuracy of the IRT parameter estimation. While our initial parameter estimation uses data from multiple question types (including short-answer and multiple-choice questions), the subsequent analysis exploring the relationship between IWFs and IRT parameters focuses solely on the 7,126 MCQs.

\subsection{Item-Writing Flaws Application}
\label{subsec:iwf_application}

\input{Table_2}

We evaluate the quality of MCQs based on the 19-criteria IWF rubric by \citet{tarrant2006frequency} (Table~\ref{table_2}). This learning science rubric is domain independent, and prior research has validated its utility in medical education, mathematics, humanities, and other STEM domains like chemistry and computer science~\citep{haladyna2002review, tarrant2006frequency, costello2018future}. Given the immense resources required for domain experts to annotate the 7,126 MCQs in our dataset, we utilize the Scalable Automatic Question Usability Evaluation Toolkit (SAQUET)~\citep{Moore2024:Automatic}, an open-source method\footnote{\url{https://github.com/StevenJamesMoore/SAQUET}} that facilitates the automated application of the IWF rubric.

SAQUET has been extensively validated across multiple educational domains, demonstrating strong alignment with expert human evaluations. In validation studies across chemistry, statistics, computer science, humanities, and healthcare domains, SAQUET achieved an overall classification accuracy of 94.13\% when evaluated against human expert annotations on 5,149 individual flaw classifications across 271 MCQs~\citep{Moore2024:Automatic}. The system demonstrated an exact match ratio of 38\% for complete question evaluations (requiring perfect classification across all 19 criteria) and a Hamming Loss of 5.9\%, indicating minimal misclassification rates. Importantly, SAQUET tends to be more conservative than human evaluators, identifying flaws more frequently (M=1.75, SD=1.26 flaws per question) compared to human assessments (M=1.31, SD=1.11), thus erring on the side of caution in question quality assessment.

Performance varied across individual IWF criteria, with surface-level flaws showing particularly strong detection rates. For example, criteria such as ``none-of-the-above'', ``all of the above'', and ``true or false'' achieved F1-scores ranging from 0.91 to 1.00 across domains. More complex pedagogical criteria showed moderate performance, with ``implausible distractors'' achieving F1-scores around 0.86 depending on domain. The micro-averaged F1 scores across all criteria ranged from 0.59 to 0.67 across the five validation domains, indicating consistent performance across different subject areas. The output of SAQUET is a labeled dataset in which each item (MCQ) is annotated with a vector $\bm{x}_i \in \{0, 1\}^{19}$ of binary indicators, specifying the presence or absence of a specific flaw as characterized by the 19-criteria rubric.

\subsection{SAQUET Technical Implementation and Domain-Specific Reliability}
\label{subsec:saquet_technical}

SAQUET employs a hybrid approach combining rule-based methods with selective LLM verification to detect the 19 IWF criteria. The open-source system categorizes flaw detection into three technical approaches based on the nature of each criterion~\citep{Moore2024:Automatic}.

\textbf{Rule-based Detection:} Eight criteria rely primarily on pattern matching and string-based techniques, including ``none-of-the-above'', ``all of the above'', ``fill-in-the-blank'', ``true or false'', ``longest answer correct'', ``negatively worded'', ``lost sequence'', and ``vague terms''. These criteria showed the highest reliability, with most achieving perfect or near-perfect F1-scores (0.91-1.00) across validation domains.

\textbf{NLP-Enhanced Detection:} Five criteria utilize foundational natural language processing techniques, including word embeddings, Named Entity Recognition (NER), and Transformer models (RoBERTa). These criteria: ``implausible distractor'', ``word repeats'', ``logical cues'', ``ambiguous information'', and ``grammatical cues'' showed moderate performance with F1-scores of up to 0.96, depending on domain and criterion complexity.

\textbf{LLM-Verified Detection:} Six criteria incorporate GPT-4o API calls for final verification, including ``absolute terms, ``more than one correct'', ``complex or k-type'', ``gratuitous information'', ``unfocused stem'', and ``convergence cues''. These criteria exhibited the most variable performance, achieving F1-scores of up to 0.89, with ``more than one correct'' proving particularly challenging due to the LLM's difficulty in reliably determining single correct answers.

\textbf{Domain-Specific Performance:} SAQUET's performance showed consistency across educational domains in the validation study. Micro-averaged F1-scores were: Chemistry (0.59), Statistics (0.65), Computer Science (0.62), Humanities (0.66), and Healthcare (0.67). When categorizing questions as acceptable (0-1 flaws) versus unacceptable (2+ flaws), SAQUET achieved a 75.3\% match rate with human evaluators across all domains.

\textbf{Validation for Current Study:} For the present study, no additional human validation of SAQUET outputs was performed on the CK-12 dataset, given the extensive prior validation and the scale of our dataset (7,126 MCQs). However, we acknowledge this as a limitation and recommend that future applications consider domain-specific validation, particularly for subject areas not represented in SAQUET's original validation study (Chemistry, Statistics, Computer Science, Humanities, Healthcare). Our dataset includes physical sciences, mathematics, and life/earth sciences, which partially overlap with but extend beyond SAQUET's validated domains.

\subsection{Analysis Methodology}
\label{subsec:analysis_methodology}

After using the student data for IRT parameter estimation and SAQUET for IWF rubric application, we define our analysis dataset as $D = \{(\alpha_i, \delta_i, \bm{x}_i)\}_{i=1}^{7126}$. Here, each MCQ $i$ in the content pool is characterized by a tuple describing its discrimination parameter $\alpha_i$, difficulty parameter $\delta_i$, and a binary vector $\bm{x}_i \in \{0, 1\}^{19}$ indicating which flaws apply. We further define domain-specific datasets ($D_{\text{Life/Earth}}$, $D_{\text{Physical}}$, $D_{\text{Math}}$) to study potential differences across subject areas (Table~\ref{table_1}). Using these datasets, we address our research questions through a mixed methodology that combines traditional regression analysis with modern machine learning algorithms.

For RQ1, we employ correlation analysis to examine how the frequency of IWFs present in MCQ's relates to their IRT difficulty and discrimination parameters. We compute Pearson’s and Spearman’s correlation coefficients to assess linear and monotonic associations, respectively, between IWF frequency and IRT parameters across each subject domain. Assumptions for Pearson’s correlation were checked visually and through descriptive statistics. Spearman’s results were emphasized as a robust complement to Pearson’s coefficients. By computing correlation coefficients models separately for the full dataset and each domain-specific subset, we investigate how flaw prevalence relates to difficulty and discrimination across subjectdomains. All analyses were conducted using the Python package SciPy \citep{Virtanen2020:Scipy}. Because difficulty and discrimination are conceptually distinct psychometric constructs, we treated them as two separate families of hypotheses. Consequently, Holm-Bonferroni adjustments were applied within each family to control the family-wise error rate (FWER) at $\alpha = 0.05$~\citep{Holm1979:Simple}.

For RQ2, we use multiple regression analysis implemented with the Python package statsmodels~\citep{Seabold2010:Statsmodels} to identify which IWF criteria are most strongly associated with MCQ difficulty and discrimination. Specifically, we fit two regression models:
\begin{align}
    \delta_i &= \beta_0 + \sum_{f \in \{1, \dots, 19\}}\beta_f x_{if} + \epsilon_i\\
    \alpha_i &= \gamma_0 + \sum_{f \in \{1, \dots, 19\}}\gamma_f x_{if} + \eta_i
\end{align}
where $\delta_i$ is difficulty, $\alpha_i$ is discrimination, and $x_{if}$ indicates the presence of flaw $f$ in MCQ $i$. The coefficients $\beta_f$ and $\gamma_f$ quantify the relationship between each IWF criteria and the difficulty and discrimination parameters, respectively. The error terms $\epsilon_i$ and $\eta_i$ capture the residual variance--that is, the portion of variability in item difficulty and discrimination that remains unexplained by the 19 IWF predictors. By estimating these models, we examine the extent to which each IWF contributes to variations in difficulty and discrimination across datasets. Analogous to RQ1, we employ Holm-Bonferroni to control the FWER at $\alpha = 0.05$.

Several assumption checks were conducted to confirm the validity of the multiple regression analyses. Multicollinearity was evaluated using the variance inflation factor (VIF), with values below 5 considered acceptable. Autocorrelation was assessed using the Durbin-Watson statistic, with values close to 2 indicating minimal autocorrelation~\cite{Field2024:Discovering}. Linearity was deemed satisfied as all predictors (i.e., the 19 IWF features) were binary categorical variables. Although the Shapiro-Wilk test indicated deviations from normality, given the large sample sizes ($N > 1000$), the regression analyses were considered robust to this violation. To address heteroscedasticity identified by the Breusch-Pagan test, HC3 robust standard errors were utilized for estimating the regression coefficients~\cite{Mackinnon1985:Some}.

For RQ3, we investigate the extent to which IWF rubric-based evaluations, derived solely from item text, can serve as a proxy for traditional validation methods that require student response data to estimate IRT parameters. Specifically, we assess the predictive power of the flaw indicator vector $\bm{x}_i$ in two tasks: (i) predicting an item's difficulty ($\delta_i$) and discrimination ($\alpha_i$) parameters; and (ii) predicting items with low discrimination ($\alpha_i < 0.5$), low difficulty ($\delta_i < -2$), and high difficulty ($\delta_i > 2$), using thresholds based on guidelines from standard IRT references~\citep{Hambleton1991:Fundamentals, Baker2001:Basics}. To this end, we train machine learning models to determine whether rubric-based flaw annotations provide sufficient predictive power to support automated item pre-screening across educational domains. We do not train models for identifying high-discrimination items, as high discrimination is typically desirable and not considered an item flaw.

Related evaluations consider a range of parametric and non-parametric machine learning algorithms, including linear/logistic regression, random forest, gradient boosting, and multi-layer perceptron (MLP), using implementations from the Python package scikit-learn~\citep{Pedregosa2011:Scikit}. For the regression tasks, we evaluate model fit using root mean squared error (RMSE) and assess predictive power using explained variance (R\textsuperscript{2}) and Pearson correlation ($r$). For the classification tasks, we measure performance using accuracy (ACC), area under the curve (AUC), and F1-score. Given the class imbalance, where approximately 90\% of items exhibit ``benign'' IRT parameter values, AUC and F1-score are particularly relevant, as they provide a robust evaluation of model performance in imbalanced classification tasks.

Our results report average performance metrics across a 5-fold cross-validation. In each fold, 80\% of the items in the dataset are used for model training and grid search-based hyperparameter selection, and 20\% are used for testing. Thus, all results are based solely on predictions for items that were not observed during training. Table~\ref{table_3} outlines the hyperparameter spaces considered for each algorithm.

\input{Table_3}


\section{Results}
\label{sec:results}

\begin{figure}[h]
\centering
\includegraphics[width=\columnwidth]{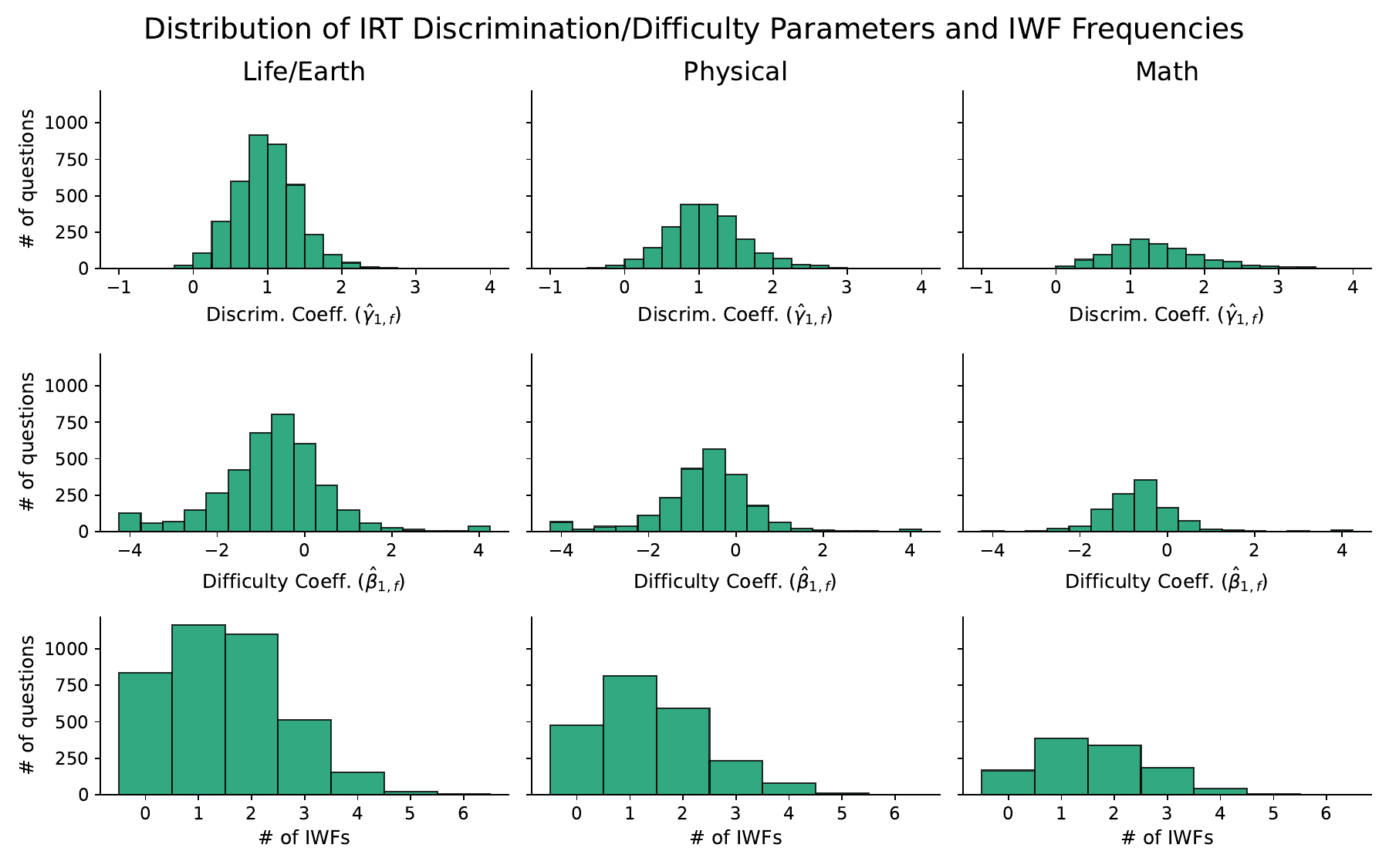}
\caption{Histograms showing the distribution of IRT discrimination and difficulty parameters, and the number of identified IWFs per question across the domain-specific datasets.}
\label{figure_1}
\end{figure}

Using data from 448,000 students, we fitted IRT models for each of the 1,033 concepts (Figure \ref{figure_1}). Assessing discrimination and difficulty parameters of all 7,126 MCQs, we flagged 789 (11.1\%) for low discrimination, 773 (10.8\%) for low difficulty, and 134 (1.9\%) for high difficulty (Table~\ref{table_4}). Across the domain-specific datasets, we observed that Life/Earth Science and Math showed the highest and lowest proportions of flagged questions, respectively (low discrimination 12.1\% vs. 7.3\%, low difficulty 10.5\% vs. 4.3\%, and high difficulty 7.3\% vs. 1.6\%). These findings suggest that the Math MCQs within individual concepts have more homogeneous difficulty levels compared to the science MCQs. Overall, we find that the vast majority of MCQs exhibit desirable IRT parameters. This implies that our item screening models have to manage class imbalance when trying to predict whether an item has desirable difficulty and discrimination (RQ3). 

In the IWF application, we found that most questions had either no flaws or very few, with 82.5\% containing at most two (Figure \ref{figure_1}), despite SAQUET leaning on the stricter side of tagging IWFs. Among the three domains, Life/Earth Science featured the highest proportion of flawless MCQs at 22.0\%. Math exhibited the highest average number of IWFs per question at 1.62. Still, all three domains demonstrated a similar distribution of IWF numbers, with an overall average of 1.48 IWFs per question. Additional details are provided in Table~\ref{table_4}, which highlights the prevalence of the five most common IWFs within each domain. The most frequent flaw identified across all domains was ``ambiguous/unclear language'' in the question text or answer options, affecting 31.3\% of all MCQs. We found ``fill-in-the-blank'' (fitb) and ``none-of-the-above'' (nota) formulations to be more prevalent in the Life/Earth (29.2\% fitb, 15.9\% nota) and Physical Science (18.4\% fitb, 10.8\% nota) domains, compared to Math (7.8\% fitb, 4.1\% nota). The ``lost-sequence'' flaw, which indicates that answer options break chronological or numerical order, was significantly more common in Math MCQs at 28.1\%. We continue by assessing the impact of IWF numbers (RQ1) and specific IWF criteria on IRT difficulty and discrimination (RQ2).

\input{Table_4}

\subsection{RQ1: Total IWF Count Correlating with IRT}
\label{subsec:rq1_results}

We conducted correlation analyses to examine how the number of IWFs relates to IRT discrimination and difficulty parameters across aggregated and domain-specific datasets (Table \ref{table_5}). First, focusing on the aggregated dataset containing all 7,126 MCQs, we observed a significant negative relationship between IWF frequencies and discrimination parameters (Pearson $r = -0.17$, $p < .001$; Spearman $r = -0.18$, $p < .001$), indicating that items with higher discrimination were less likely to contain IWFs. This pattern was consistent across Life/Earth (Pearson $r = -0.20$, $p < .001$; Spearman $r = -0.20$, $p < .001$) and Physical Science (Pearson $r = -0.28$, $p < .001$; Spearman $r = -0.28$, $p < .001$), suggesting that well-discriminating items in these domains were generally written with fewer flaws. The relationship between IWF frequencies and difficulty parameters showed mixed results. In Life/Earth, the domain with the most questions, there was a weak but significant negative association (Pearson $r = -0.08$, $p < .001$; Spearman $r = -0.05$, $p < .05$), indicating that easier items were more prone to contain flaws. However, in Physical Science and Math, we did not find clear relationships between difficulty parameters and IWF frequencies. These findings show the importance of analyzing specific types of IWFs, as some may be linked to higher item difficulty while others correspond to lower difficulty.

\input{Table_5}

\subsection{RQ2: Identifying High-Impact IWFs}
\label{subsec:rq2_results}

Next, we explore how different writing flaws can differentially affect the cognitive demands of test items beyond what is revealed by overall IWF frequency.The multiple regression analysis aimed to identify which specific IWF criteria are most strongly associated with MCQ's discrimination and difficulty parameters. Diagnostic checks confirmed that assumptions for analyses were satisfied: The maximum VIF observed across the four datasets was 1.63, indicating no concerns regarding multicollinearity. The Durbin-Watson statistics ranged from 1.56 to 1.96, suggesting negligible autocorrelation among residuals. Although the Shapiro-Wilk tests revealed deviations from normality, the large sample sizes ($N > 1000$) justified robustness against this violation. Robust HC3 standard errors were utilized to account for heteroscedasticity detected by the Breusch-Pagan test, ensuring valid inference~\cite{Fox2015:Applied}.

\begin{figure*}[h]
    \centering
    \includegraphics[width=\linewidth]{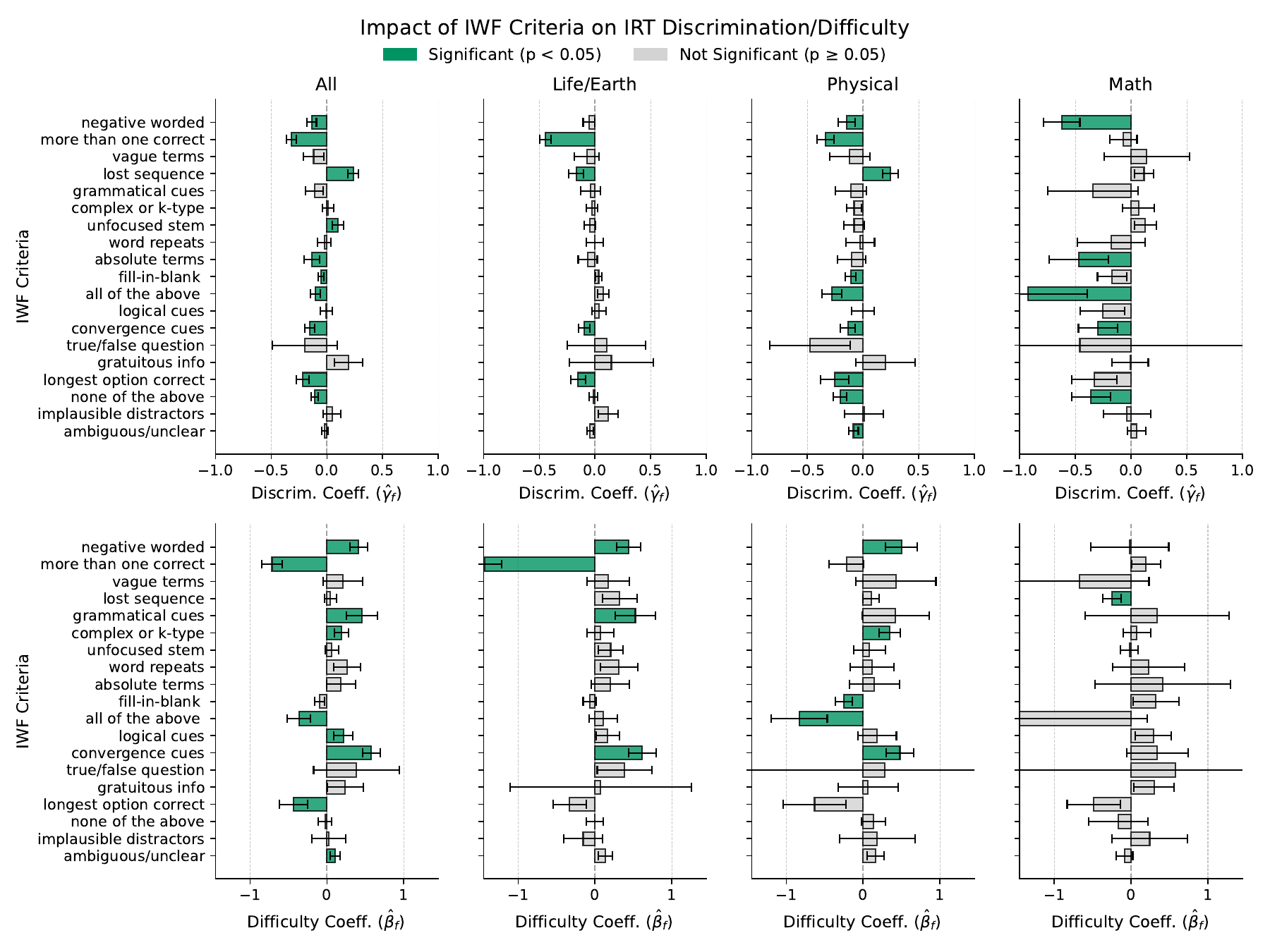}
    \caption{Multiple‐regression analysis examining the strength of association between each IWF criterion and IRT discrimination and difficulty parameters across the domain-specific datasets. The figure displays estimated coefficients with 95\% confidence intervals. Relationships that remain statistically significant after controlling the family-wise error rate at $\alpha = 0.05$ using the Holm-Bonferroni method are highlighted in green.} 
    \label{figure_2}
\end{figure*}

For each dataset and IWF criterion $f \in \{1, \dots, 19\}$, Figure~\ref{figure_2} presents the estimated discrimination ($\hat{\gamma}_f$) and difficulty coefficients ($\hat{\beta}_f$), along with their 95\% confidence intervals. Statistically significant coefficients ($p < 0.05$) are highlighted in green. Examining the combined dataset of 7,126 MCQs, we found significant associations between IRT discrimination and 10 of the 19 IWF criteria, while 9 criteria were significantly associated with difficulty. Among the domain-specific datasets, Physical Science exhibited the highest number of significant discrimination coefficients (9), suggesting stronger associations with diverse IWF criteria compared to Life/Earth Science (4) and Math (5). Difficulty coefficients were more frequently significant for Physical Science (5) and Life/Earth Science (10). In Math, ``lost sequence'' was the only criterion for which our FWER-controlled approach could establish a significant association with MCQ difficulty. 

When examining individual IWFs, we found that the flaws most negatively associated with IRT discrimination and difficulty parameters were ``more than one correct'' ($\hat{\gamma}_f = -0.317$, $\hat{\beta}_f = -0.715$), ``longest option correct'' ($\hat{\gamma}_f = -0.216$, $\hat{\beta}_f = -0.437$), and ``all of the above'' ($\hat{\gamma}_f = -0.101$, $\hat{\beta}_f = -0.364$). These flaws likely introduce textual cues that inadvertently signal the correct answer, thereby diminishing item quality. In contrast, the ``lost sequence'' flaw had a positive discrimination coefficient ($\hat{\gamma}_f = 0.314$). Interestingly, this flaw was associated with higher discrimination in Life/Earth Science but lower discrimination in Physical Science, suggesting that the same flaw may have differing effects across domains. In Mathematics, IWFs showed the strongest associations with reduced discrimination, particularly ``all of the above'' ($\hat{\gamma}_f = -0.926$), ``negative wording'' ($\hat{\gamma}_f = -0.622$), and ``absolute terms'' ($\hat{\gamma}_f = -0.470$). Several IWFs were also linked to increased  difficulty, including ``convergence cues'' ($\hat{\beta}_f = 0.579$), ``grammatical cues'' ($\hat{\beta}_f = 0.455$) and ``negative wording'' ($\hat{\beta}_f = 0.413$). These flaws may increase cognitive load or introduce confusion unrelated to content knowledge.

\subsection{RQ3: IWF-Based IRT Predictions}
\label{subsec:rq3_results}

Using the IWF annotations as input features, we trained machine learning models to predict questions' difficulty and discrimination parameters. The performance of the resulting models, as shown in Table~\ref{table_6}, varied across educational domains and predicted parameters. For the discrimination parameter, when trained on the dataset comprising all 7,126 MCQs, the models achieved Pearson correlation coefficients ($r$) ranging from 0.348 to 0.377 and explained variance ($R^2$) ranging from 0.121 to 0.141, indicating moderate predictive strength. For the difficulty parameter, the Random Forest ($r=0.457$, $R^2=0.209$) and MLP ($r=0.465$, $R^2=0.216$) models showed the highest Pearson correlations and explained variance, suggesting more effective utilization of the IWF features. Notably, non-linear models (Random Forest, Gradient Boosting, and MLP) consistently outperformed the linear regression model, indicating that modeling non-linear interactions between IWF features can improve predictive accuracy. We further observed substantial differences between the domain-specific models. For instance, in Life/Earth Science, the Random Forest model achieved the highest Pearson correlation ($r=0.510$) and explained variance ($R^2=0.259$). However, in Math, all models struggled with both discrimination (max $R^2=0.071$) and difficulty predictions (max $R^2=0.030$).

\input{Table_6}

We evaluate the utility of IWF features for predicting MCQs with low discrimination and low/high difficulty in Life/Earth and Physical Science datasets, where the prior analysis confirmed the predictive power of IWFs. Table~\ref{table_7} shows that models trained on IWF features achieve AUC scores of up to 0.746 (random forest) and 0.799 (gradient boosting) for low discrimination, and 0.825 (gradient boosting) and 0.784 (logistic regression) for low difficulty. While the AUC scores suggest strong predictive performance, F1 scores remain relatively low for low-discrimination MCQs (peaking at 0.364 for Life/Earth and 0.435 for Physical Science), indicating potentil challenges due to class imbalance (Table~\ref{table_4}). In contrast, F1 scores for low-difficulty MCQs are considerably higher, with logistic regression achieving 0.649 for Life/Earth and 0.516 for Physical Science, suggesting that IWFs are informative for identifying low-difficulty MCQs. In contrast, none of the models trained to predict high difficulty MCQs outperformed a baseline that always predicts the majority class. This is likely due to class imbalance and the limitation that IWFs, which focus on linguistic features, are not are not designed to assess the knowledge required to answer domain-specific questions.

\input{Table_7}

\begin{figure}[h]
    \centering
    \includegraphics[width=0.7\columnwidth]{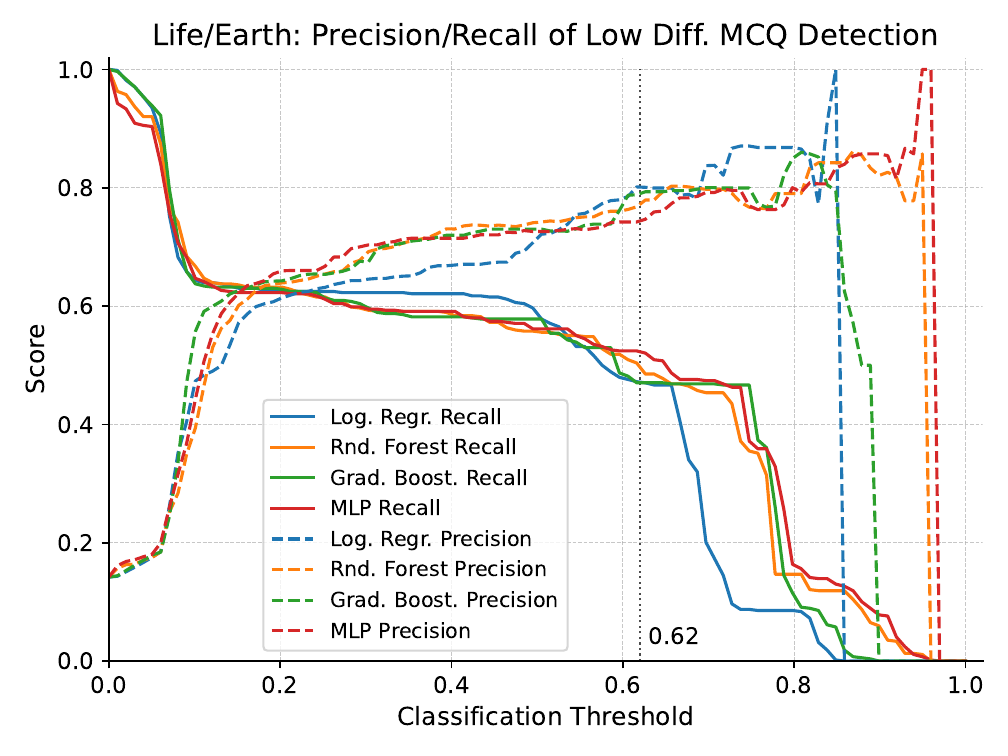}
    \caption{Precision and recall curves for predicting low-difficulty Life/Earth Science MCQs for different classifiers. The curves show trade-off between recall (solid lines) and precision (dashed lines) across classification thresholds. Using a threshold of 0.62, logistic regression achieves a precision of 0.801 and a recall of 0.472.}
    \label{figure_3}
\end{figure}

Since the IWF-based classification models demonstrated the highest predictive performance for identifying low-difficulty MCQs in the Life/Earth Science dataset, we conducted a follow-up analysis to assess their potential for automated item pre-screening. Figure~\ref{figure_3} illustrates the trade-offs between precision and recall across different classification thresholds. Recall represents the proportion of low-difficulty MCQs correctly identified by the models, while precision reflects the fraction of flagged MCQs that genuinely belong to the low-difficulty category. By setting the classification threshold to 0.62, the logistic regression model achieves a high precision of 0.801 while maintaining a moderate recall of 0.472. This balance demonstrates the practical utility of IWF-based classifiers in supporting experts in test item development by enabling early identification of low-difficulty questions, potentially lowering the need for student data collection.

\section{Discussion}
\label{sec:discussion}

Our study integrated statistical and machine learning methodologies with large-scale student data, capturing interactions with thousands of questions. This approach provided insights into how qualitative aspects of question design influence traditional measures of question performance derived from IRT~\citep{Ayala2013:Theory}. Specifically, we examined the relationships between the standard 19-criteria IWF rubric~\citep{tarrant2006frequency} for MCQs and IRT parameters across various educational domains (physical science, mathematics, and life/earth sciences). Our findings offer quantitative evidence demonstrating how the frequency and specific types of IWFs impact question discrimination and difficulty. Additionally, we validated the utility of IWF evaluations as features for predicting IRT parameters and for identifying low-difficulty questions during test development.

Across the three domains we examined, the frequency of IWFs shows a clear relationship with item discrimination, but its relationship to item difficulty is more domain-specific. In the science domains, MCQs with fewer flaws showed significantly higher discrimination, indicating that IWFs can diminish a question’s reliability. However, this relationship did not hold for Math, where no significant correlation was found. This may be because the nature of flaws differs by domain, such as the ``lost sequence`` flaw being common in Math MCQs, but having a negligible effect on discrimination, unlike the flaws more prevalent in the science domains. As noted in prior work, certain IWFs may inadvertently aid students in guessing the correct answer, such as ``longest answer correct'' or ``all of the above'' \citep{tarrant2006frequency,haladyna2002review}, while others add confusion unrelated to the content itself, like ``negative wording''. In either case, such flaws can distort how accurately the question discriminates between more- and less-knowledgeable students. The link to item difficulty was less consistent, as in Life/Earth Science, we observed that easier items contained more flaws, suggesting the frequent presence of the flaws that may effectively simplify questions. This trend did not surface in Physical Science or Math, suggesting the possibility that IWFs and item difficulty interact in a domain-dependent manner. It may also reflect variations in how effectively the automated IWF detection methods operate across different subject areas, aligning with previous findings that showed stronger performance in Humanities and Healthcare than in Chemistry~\citep{Moore2024:Automatic}. Consequently, while IWF frequency appears to be a reliable indicator of item discrimination overall, its utility in predicting item difficulty likely hinges on both the domain and the strengths or limitations of automated detection techniques. These domain-specific patterns suggest that practical implementations should allow test developers to customize IWF screening based on subject area, potentially using different weighting schemes or thresholds for different educational domains.

Our findings indicate that most IWF criteria can significantly influence both item discrimination and difficulty, though certain flaws exhibit particularly strong and consistent effects. Specifically, the three flaws of ``more than one correct'', ``longest option correct'', and ``all of the above'' show the most substantial \textit{negative associations} across both parameters. This is likely because they introduce cues that enable students to guess the correct answer without engaging with the intended knowledge. Additionally, flaws such as ``convergence cues'', ``grammatical cues'', and ``negative wording'' were associated with higher item difficulty, but worse discrimination. This suggests that these flaws may elevate cognitive load by requiring students to navigate complex text structures rather than directly demonstrating their domain knowledge. Consistent with previous work, the presence of ``all of the above'' as an answer choice and the ``longest option correct'' decreased the difficulty of the question, while also weakening its discrimination~\citep{rush2016impact}. While some flaws consistently diminish both question quality and rigor, our findings highlight how specific IWFs exert their influence differently. They might make questions easier to guess or introduce additional cognitive demands that may confuse students. Others appear to have more nuanced effects, warranting further investigation into their role in shaping assessment validity and fairness. The strength of these associations, particularly for item difficulty, was more pronounced in the science domains of Physical and Life/Earth compared to Math. This is likely because mathematical difficulty is driven more by the inherent conceptual complexity and procedural logic of the problem rather than the textual construction of the question. In science domains, ambiguous language or confusing textual cues may have a greater impact on an item's perceived difficulty, whereas in Math, these structural flaws are often secondary to the cognitive demands of the underlying reasoning task, which the IWF rubric is not designed to assess.

Machine learning models trained to predict question discrimination and difficulty parameters based on IWF annotations achieve moderate predictive power, with performance varying across subject domains (Table~\ref{table_7}). Notably, prediction accuracy was higher for Life/Earth and Physical Science questions compared to Math, particularly for difficulty estimation. This aligns with our regression analysis, which revealed stronger associations between IWFs and difficulty parameters in the Science domains (Table~\ref{figure_2}). Across all prediction tasks, non-linear models, such as gradient boosting and MLP, consistently outperformed linear models, demonstrating the need to capture complex interactions between individual IWF features. Additionally, this limited performance is expected as IWF features are designed to be domain-general. By focusing on an item's textual structure, they deliberately ignore the primary drivers of its psychometric properties: the domain-specific knowledge and cognitive complexity required to answer it. While a flawed question is often a poor question, a well-written one is not necessarily easy or difficult.

To assess the practical utility of IWF features for item screening, we evaluated classification models designed to identify questions with low discrimination, low difficulty, and high difficulty. Our results suggest that by selecting a classification threshold that balances precision and recall, IWF-based models can assist domain experts in identifying low-difficulty questions early. In particular, criteria such as ``all of the above'' and ``longest option correct'' showed strong associations with low item difficulty, likely explaining why our classifiers performed significantly better at predicting low-difficulty MCQs compared to high-difficulty ones. The latter task likely requires models to assess the specific knowledge needed to answer a question within a given domain, showing a potential limitation of IWF features for difficulty prediction. From a practical standpoint, these findings suggest that IWF-based screening could be most effectively integrated early in the test development cycle, where content experts could use automated flagging to prioritize which items require closer review before pilot testing. Test development workflows could incorporate IWF analysis as a first-pass quality check, allowing developers to address potential issues with discrimination and difficulty before expensive data collection phases. This approach would be particularly valuable for identifying low difficulty items that may need revision, given our models' strong performance in this classification task.

By examining how qualitative question design guidelines~\citep{tarrant2006frequency} align with robust statistical measurements derived from large student datasets, our study contributes to ongoing research efforts on characterizing effective instructional design principles~\citep{Koedinger2013:Instructional}. From a learning science perspective, rubrics serve as distilled representations of expert knowledge used to assess the quality of educational materials and instruction ~\citep{Price2019:Comparison, Watson2021:Alignment,Liu2024:Assessing, Aher2025:AI}. Understanding the relationship between expert evaluation rubrics and student learning is crucial, especially as AI-driven learning technologies increasingly rely on textual descriptions of effective pedagogical strategies~\citep{Sonkar2023:Class, Schmucker2024:Ruffle, Puech2024:Towards, Jurenka2024:Towards}. In practice, IWF-based screening tools could serve as automated quality assurance systems within existing test development platforms, flagging items for expert review and suggesting specific revision targets based on detected flaws. Such integration would allow test developers to maintain human oversight while benefiting from systematic, scalable item screening.

\subsection{Equity and Ethical Implications}
While our findings demonstrate the predictive utility of IWFs for identifying problematic MCQs, they also raise important concerns about equity and fairness in automated assessment validation. Certain IWFs that significantly impact item difficulty, such as ``ambiguous/unclear language'', ``grammatical cues'', and ``negative wording'', may disproportionately disadvantage specific student populations. For instance, students whose native language is not English may face additional cognitive burden when encountering questions with grammatical inconsistencies or complex linguistic structures, potentially masking their true subject matter knowledge. Similarly, students from diverse cultural or socioeconomic backgrounds might interpret ambiguous language differently, leading to systematic differences in performance that reflect language proficiency rather than domain expertise. Our finding that negative wording increases item difficulty suggests that such constructions may introduce barriers unrelated to the content the question is assessing, potentially creating inequitable testing conditions for students with varying linguistic backgrounds or cognitive processing styles~\citep{cormier2022linguistic}.

The automation of item validation through IWF detection introduces ethical considerations that extend beyond individual question quality. While these tools can efficiently flag potentially problematic items, they risk reinforcing existing biases embedded in both the training data and the assessment practices from which the IWF criteria were originally derived~\cite{Kizilcec2022:Algorithmic}. If automated systems consistently flag certain types of language or question structures that are more familiar to specific demographic groups, they may inadvertently perpetuate systemic inequities in educational assessment. Furthermore, over-reliance on automated validation could diminish the role of human expertise in recognizing context-specific cultural or linguistic sensitivities that algorithmic approaches might miss. To mitigate these risks, we recommend that IWF-based screening tools be implemented as decision-support systems rather than replacement technologies, maintaining human oversight in the item validation process. In this workflow, the flawed questions can be remedied to provide a more accurate and true assessment of student learning, while also leading to more accurate IRT parameters.

\section{Limitations and Future Work}

While this study established relationships between a domain-general IWF rubric~\citep{tarrant2006frequency} and statistical measures of question difficulty and discrimination derived from IRT~\citep{Ayala2013:Theory}, several limitations should be considered. Our analysis was conducted entirely within a single large-scale online tutoring platform, which may limit the generalizability of our findings across different educational contexts. The student population, instructional approaches, and assessment practices within this platform may not be representative of traditional classroom settings, standardized testing environments, or other online learning systems. Additionally, our analysis was limited to science and mathematics courses at the middle and high school levels, although our dataset comprised over 7,000 questions and hundreds of thousands of student responses to provide strong internal validity. The relationships between IWFs and IRT parameters observed in this specific educational technology context may not hold across different delivery modalities, student demographics, or institutional settings. Future research should validate these findings across multiple platforms, educational institutions, and delivery formats to establish broader applicability. Further exploration is also needed across other subject areas, including language, humanities, and social sciences, as well as in higher and professional education contexts, particularly in medical education, where MCQ-based assessments are widely used~\citep{downing2005effects, rush2016impact}

Although the IWF rubric provides a domain-general methodology for experts to assess the pedagogical soundness of test items without relying on student data~\citep{tarrant2006frequency}, our findings provide the first quantitative evidence that IWF features are statistically significant predictors of MCQ difficulty and discrimination (Table~\ref{table_7}). The predictive signal is moderate, which is expected, as IWFs focus on broad design principles, such as ensuring that all distractors are plausible, but do not capture domain-specific nuances related to the knowledge required to solve a particular test item. Our reliance on automated IWF detection through SAQUET introduces additional limitations that may affect the validity of our findings. While SAQUET has demonstrated high accuracy in identifying common item-writing flaws~\citep{Moore2024:Automatic}, automated tools may miss subtle contextual nuances that human experts would readily identify, or conversely, flag false positives where apparent flaws do not actually impact item quality. The tool's performance may also vary across different domains, as evidenced by previous research showing stronger performance in some subject areas than others. Additionally, automated detection systems are inherently limited by their training data and may systematically under- or over-detect certain types of flaws, potentially biasing our analysis of IWF-IRT relationships.

An item may fully adhere to IWF guidelines yet still exhibit high or low difficulty levels depending on the complexity of the subject knowledge it assesses. To address these limitations, future research could explore hybrid approaches that combine the interpretability of IWF-based evaluations with the predictive power of deep learning models, which estimate IRT parameters based on semantic analyses of question text~\citep{Alkhuzaey2024:Text}. Another promising direction is the development of enhanced evaluation rubrics that integrate human domain expertise with data-driven insights generated by machine learning algorithms, thereby improving their predictive accuracy~\citep{Koh2020:Concept, Ludan2023:Interpretable, Barany2024:Chatgpt}.

Across the courses examined in this study, the IWF analysis identified an average of 1.48 writing flaws per MCQ (Table~\ref{table_4}). While many of these flaws may have minimal impact on student learning outcomes, addressing them remains essential for ensuring content quality. Future work will focus on developing AI-assisted content authoring tools to support domain experts in MCQ generation and refinement~\citep{Moore2024:Automatic}. Recent advancements in LLM-enabled pipelines for question generation and validation offer promising directions~\citep{Bhandari2024:Evaluating, Liu2024:Leveraging, He2024:Evaluating}. To enhance the efficiency of question validation, future research will explore natural language processing and reinforcement learning to reduce the amount of student response data required for reliable IRT parameter identification~\citep{Yancey2024:Bert, sharpnack2024autoirt, Sharpnack2024:Banditcat}.
 
Lastly, we emphasize the broader utility of evaluation methodologies that integrate generative AI to scale qualitative assessments based on learning science rubrics with statistical measures derived from student data. This hybrid approach can generate robust and actionable insights for improving educational practice. Future research will extend this framework to evaluate other types of educational materials, such as hints~\citep{Price2019:Comparison, Schmucker23:Learning, Stamper2024:Enhancing}, textbooks~\citep{Watson2021:Alignment}, and illustrations~\citep{Angra2018:Graph}. Additional directions include examining the predictive validity of rubric-based evaluations in educational domains such as project-based learning~\citep{Catete2016:Developing, Goyal2022:Meta, Aher2025:AI}, discourse analysis~\citep{Liu2024:Assessing, Borchers2024:Combining} and programming education~\citep{Catete2016:Developing, Saito2021:Validation}.

\section{Conclusion}
\label{sec:conclusion}

This paper explored relationships between the 19-criteria Item-Writing Flaws (IWFs) rubric, a domain-general qualitative method for text-based question validation~\citep{tarrant2006frequency}, and item response theory (IRT), a traditional, student data-driven approach to assessing question quality~\citep{Ayala2013:Theory}. Using an automated method, we applied the rubric to 7,126 multiple-choice questions spanning mathematics, physical, and life/earth science domains, analyzing how the number and types of IWFs impact question difficulty and discrimination parameters. Three key findings emerged. First, a higher number of IWFs was associated with lower item difficulty and discrimination in life/earth sciences, while the relationship was less consistent in mathematics and physical sciences. Second, specific IWF criteria strongly correlated with question difficulty, such as “longest option correct” for easier items and “convergence cues” for harder ones, demonstrating how superficial textual cues can compromise an otherwise well-designed question. Third, while models trained on IWF features did not match the precision of IRT-based methods, they showed promise for preliminary screening, particularly in identifying low-difficulty questions.

These findings show the dual role of domain-agnostic and domain-specific factors in developing high-quality test items. On the one hand, a rubric that flags generic writing flaws can serve as a scalable ``first pass'', helping content authors identify potential design issues before pilot testing. On the other hand, IWF features alone are only moderate predictors of IRT parameters, with predictive strength varying across educational domains. This highlights that IWF-based evaluation cannot replace traditional student data-dependent methods, such as those embodied in IRT. Future work could explore hybrid approaches that integrate the interpretability of human-readable rubrics with the flexibility of machine-learning models capable of capturing semantic information related to domain-specific knowledge to enhance the accuracy of IRT parameter predictions. This systematic alignment of qualitative rubrics with quantitative validation not only helps improve item quality at scale but also ensures that computer-assisted assessments support fair, reliable, and pedagogically meaningful testing.

\section*{Acknowledgments}

We thank Microsoft for support in the form of Azure computing and access to the OpenAI API through a grant from their Accelerate Foundation Model Academic Research Program. We thank the CK-12 Foundation (ck12.org) for providing access to their learning materials and to data on student responses to those materials. This research was supported in part by the AFOSR under award FA95501710218.



\bibliographystyle{elsarticle-harv}
\bibliography{bibliography.bib}

\end{document}

%% file: Table_1.tex

\begin{table}[h]
\centering
\setlength{\tabcolsep}{8pt}
\caption{Dataset overview. The first three rows indicate the number of concepts, questions and multiple-choice questions (MCQs). The next three rows describe the number of students, practice sessions and responses providing data to estimate the IRT parameters. The last two rows show the average number of students and questions in each concept.}
\begin{tabular}{lcccc}
\hline
& \textbf{All} & \textbf{Life/Earth} & \textbf{Physical} & \textbf{Math} \\
\hline
\# of concepts & 1,033 & 563 & 336 & 134 \\
\# of questions & 13,158 & 7,212 & 4,331 & 1,649 \\
\# of MCQs & 7,126 & 3,792 & 2,206 & 1,128 \\
\hline
\# of students & 448k &  265k & 169k & 44k \\
\# of sessions & 1.9M &  1.1M & 0.6M & 0.15M \\
\# of responses & 21.1M & 12.6M & 7.0M & 1.6M \\
\hline
\# students/concepts & 1,848 & 1,983 & 1,902 & 1,155 \\
\# questions/concepts & 12.8 & 12.8 & 12.9 & 12.3 \\
\hline
\end{tabular}
\label{table_1}
\end{table}

%% file: Table_2.tex

\begin{table}
\centering
\footnotesize 
\setlength{\tabcolsep}{7pt}
\caption{Criteria of the Item-Writing Flaw (IWF) rubric~\citep{tarrant2006frequency}.}
\begin{tabularx}{\textwidth}{lX}
\hline
\textbf{IWF Criteria} & \textbf{Definition} \\
\hline
Ambiguous/Unclear & The question text and options should be written in clear, precise language to avoid confusion \\
\hline
Implausible Distractors & All incorrect answer choices should be realistic and plausible \\
\hline
None of the Above & Avoid using any variation of ``none of the above'' since it primarily tests students' ability to identify wrong answers \\
\hline
Longest Option Correct & The correct answer should not be noticeably longer or contain more detail than the other options, as this can unintentionally guide students to the correct answer \\
\hline
Gratuitous Information & Avoid unnecessary details in the question text that do not contribute to answering the question \\
\hline
True/False Question & Avoid answer choices structured as a series of true or false statements \\
\hline
Convergence Cues & Ensure answer choices do not contain overlapping words that might hint at the correct answer \\
\hline
Logical Cues & Avoid clues in the stem and the correct option that can help the test-wise student to identify the correct answer \\
\hline
All of the Above & Avoid using any variation of ``all of the above'' as students can guess the answer just by recognizing one correct option  \\
\hline
Fill in the Blank & Avoid missing words in the middle of a sentence, as this forces students to rely on partial information \\
\hline
Absolute Terms & Avoid extreme words like ``always'' or ``never'' in answer choices, as these are usually false \\
\hline
Word Repeats & Ensure words or phrases from the question text are not repeated only in the correct answer, as this can inadvertently reveal the right choice \\
\hline
Unfocused Stem & The question text should be clear and specific so students can understand it without needing to read the answer choices \\
\hline
Complex or K-type & Avoid overly complex questions that require selecting from a number of possible combinations of the responses, as they may test strategy rather than knowledge \\
\hline
Grammatical Cues & All options should be grammatically consistent with the question text and should be parallel in style and form \\
\hline
Lost Sequence & Arrange options in a logical order (e.g. chronological or numerical) to improve readability and fairness \\
\hline
Vague Terms & Avoid the use of vague words (e.g. frequently, occasionally) in the options as their meaning can be subjective \\
\hline
More than One Correct & In single-answer multiple-choice questions, ensure there is a single best answer to avoid ambiguity \\
\hline
Negative Wording & Avoid usage of negative wording in the question text, as it can confuse students \\
\hline
\end{tabularx}
\label{table_2}
\end{table}

%% file: Table_3.tex
\begin{table}[!h]
\centering
\setlength{\tabcolsep}{8pt}
\caption{Hyperparameters considered during model training.}
\begin{tabular}{ll}
    \hline
    \textbf{Model} & \textbf{Parameter} \\
    \hline
    Linear     & $\text{penalty\_weight} \in \{10^i\}_{i=-4}^4$ \\
    Regression & $\text{penalty: l2}$ \\
    \hline
    Random & $\text{n\_estimators} \in \{50, 100, 200, 300\}$ \\
    Forest & $\text{max\_depth} \in \{\text{None}, 5, 10, 20\}$ \\
                  & $\text{min\_samples\_split} \in \{2, 5, 10\}$ \\
    \hline
    Gradient & $\text{n\_estimators} \in \{50, 100, 200, 300\}$ \\
    Boosting & $\text{learning\_rate} \in \{0.001, 0.01, 0.1, 0.2, 0.3\}$ \\
                    & $\text{max\_depth} \in \{3, 5, 7, 10\}$ \\
    \hline
    Multi-layer & $\text{hidden\_layer\_sizes} \in \{10, 50, 100\}^{\{1,2\}}$ \\
    Perceptron & $\text{activation} \in \{\text{relu}, \text{tanh}\}$ \\
        & $\text{learning\_rate\_init} \in \{10^i\}_{i=-4}^0$ \\
    \hline
\end{tabular}
\label{table_3}
\end{table}

%% file: Table_4.tex

\begin{table}[h]
\centering
\setlength{\tabcolsep}{8pt}
\caption{Analysis Overview. The first section details the total number of questions and those flagged for low discrimination and low/high difficulty based on IRT analysis. The second section reports total number of IWFs identified and average per question. The last section highlights the five most common IWFs and their prevalence across domains.}
\begin{tabular}{lcccc}
\hline
 & \textbf{All}  & \textbf{Life/Earth} & \textbf{Physical} & \textbf{Math} \\
\hline
\# of questions  &  7,126 & 3,792 & 2,206 & 1,128 \\
\hspace{2pt} - low discrimination & 773 & 459 & 232 & 82 \\
\hspace{2pt} - low difficulty & 789 & 538 & 203 & 48 \\
\hspace{2pt} - high difficulty & 134 & 78 & 38 & 18 \\
\hline
\# of IWFs       & 10,537 & 5,647 & 3,062 & 1,828 \\
\# of IWFs/quest.   &  1.479 & 1.489 & 1.388 & 1.621 \\
\hline
ambiguous/unclear    & 31.3\% & 27.8\%  & 30.0\% & 45.9\% \\
fill in the blank     & 22.4\% & 29.2\%  & 18.4\% &  7.8\% \\
multiple correct  & 14.1\% & 14.3\%  & 14.4\% & 12.7\% \\
none of the above & 12.5\% & 15.9\%  & 10.8\% &  4.1\% \\
lost sequence     & 10.1\% &  2.8\%  & 13.6\% & 28.1\% \\
\hline
\end{tabular}
\label{table_4}
\end{table}



%% file: Table_5.tex
\begin{table}[h]
\centering
\setlength{\tabcolsep}{4pt}
\caption{Correlation analysis examining relationships between the number of IWFs and IRT discrimination and difficulty parameters across domains. We report estimated Pearson and Spearman coefficients, 95\% confidence intervals, and Holm-Bonferroni-adjusted $p$-values, controlling the family-wise error rate (FWER) at $\alpha = 0.05$.}
\resizebox{\linewidth}{!}{%
\begin{tabular}{lccccccccccc}
\hline
 &&& \multicolumn{4}{c}{\textbf{Discrimination}} && \multicolumn{4}{c}{\textbf{Difficulty}} \\
& $n$ && $r_{\text{Pear}}$ & $p_{\text{Pear}}$ &
   $r_{\text{Spear}}$ & $p_{\text{Spear}}$ && $r_{\text{Pear}}$ & $p_{\text{Pear}}$ & $r_{\text{Spear}}$ & $p_{\text{Spear}}$ \\
\hline
All        & 7,126 && -0.17±.02 & .000 & -0.18±.02 & .000 && -0.04±.02 & .008 & -0.01±.02 & 1.00 \\
Life/Earth & 3,792 && -0.20±.03 & .000 & -0.20±.03 & .000 && -0.08±.03 & .000 & -0.05±.03 & .015 \\
Physical   & 2,206 && -0.28±.04 & .000 & -0.28±.04 & .000 && 0.02±.04 & 1.00 & 0.06±.04 & .039 \\
Math       & 1,128 && -0.03±.06 & .786 & -0.01±.06 & .786 && 0.04±.06 & .955 & 0.00±.06 & 1.00 \\
\hline
\end{tabular}}
\label{table_5}
\end{table}

%% file: Table_6.tex
\begin{table}[h]
\centering
\caption{Regression Task. We train models that employ IWF features to predict the discrimination and difficulty parameters of MCQs. After controlling the family-wise error rate (FWER) at $\alpha = 0.05$ using the Holm–Bonferroni method, all Pearson correlation coefficients ($r$) are statistically significant at $p < 0.001$.}
\resizebox{\linewidth}{!}{%
\begin{tabular}{lccc|ccc|ccc|ccc}
\hline
& \multicolumn{3}{c}{\textbf{All}}              &  \multicolumn{3}{c}{\textbf{Life/Earth}}       &  \multicolumn{3}{c}{\textbf{Physical}}    &  \multicolumn{3}{c}{\textbf{Math}}            \\
& RMSE & $R^2$ & $r$ & RMSE & $R^2$ & $r$ & RMSE & $R^2$ & $r$ & RMSE & $R^2$ & $r$ \\
\hline

\textbf{Discrim.} &&&&&&&&&&&& \\
Lin. Regr.   & 0.491 & 0.121 & 0.348 & 0.396 & 0.129 & 0.359 & 0.476 & 0.178 & 0.422 & 0.666 & 0.071 & 0.269 \\
Rnd. Forest  & 0.487 & 0.138 & 0.373 & 0.396 & 0.131 & 0.362 & 0.475 & 0.185 & 0.430 & 0.666 & 0.071 & 0.268 \\
Grad. Boost. & 0.485 & 0.142 & 0.377 & 0.395 & 0.132 & 0.364 & 0.472 & 0.192 & 0.439 & 0.666 & 0.071 & 0.268 \\
MLP          & 0.486 & 0.141 & 0.375 & 0.396 & 0.130 & 0.361 & 0.474 & 0.187 & 0.433 & 0.666 & 0.072 & 0.273 \\
\hline
\textbf{Difficulty} &&&&&&&&&&&& \\ 
Lin. Regr.   & 1.128 & 0.115 & 0.338 & 1.161 & 0.189 & 0.435 & 1.099 & 0.102 & 0.319 & 0.914 & 0.017 & 0.141 \\
Rnd. Forest  & 1.067 & 0.209 & 0.457 & 1.109 & 0.259 & 0.510 & 1.054 & 0.174 & 0.420 & 0.917 & 0.012 & 0.156 \\
Grad. Boost. & 1.064 & 0.213 & 0.462 & 1.111 & 0.257 & 0.507 & 1.062 & 0.161 & 0.405 & 0.914 & 0.018 & 0.136 \\
MLP          & 1.062 & 0.216 & 0.465 & 1.106 & 0.264 & 0.514 & 1.043 & 0.192 & 0.438 & 0.908 & 0.030 & 0.176 \\
\hline
\end{tabular}}
\label{table_6}
\end{table}

%% file: Table_7.tex
\begin{table}
\centering
\setlength{\tabcolsep}{6pt}
\caption{In this classification task, we train models that employ IWF features to predict MCQs with low discrimination and low/high difficulty. To highlight class imbalance, we include a baseline assigning all MCQs to the majority class in gray.}
\begin{tabular}{lcccccccc}
\hline
\textbf{Task} &&  \multicolumn{3}{c}{\textbf{Life/Earth}} &&  \multicolumn{3}{c}{\textbf{Physical}} \\
&& ACC & AUC & F1 && ACC & AUC & F1 \\
\hline
\textbf{Disc. Low} && \textcolor{gray}{0.879} & \textcolor{gray}{0.500} & \textcolor{gray}{0.000} && \textcolor{gray}{0.895} & \textcolor{gray}{0.500} & \textcolor{gray}{0.000} \\
Log. Regr.   && 0.880 & 0.736 & 0.249 && 0.909 & 0.784 & 0.435 \\
Rnd. Forest  && 0.890 & 0.746 & 0.344 && 0.907 & 0.781 & 0.403 \\
Grad. Boost. && 0.888 & 0.741 & 0.354 && 0.910 & 0.799 & 0.432 \\
MLP          && 0.882 & 0.730 & 0.364 && 0.908 & 0.779 & 0.400 \\
\hline
\textbf{Diff. Low} && \textcolor{gray}{0.858} & \textcolor{gray}{0.500} & \textcolor{gray}{0.000} && \textcolor{gray}{0.908} & \textcolor{gray}{0.500} & \textcolor{gray}{0.000} \\
Log. Regr.   && 0.910 & 0.818 & 0.649 && 0.934 & 0.778 & 0.516 \\
Rnd. Forest  && 0.910 & 0.809 & 0.636 && 0.932 & 0.760 & 0.498 \\
Grad. Boost. && 0.910 & 0.825 & 0.644 && 0.933 & 0.784 & 0.506 \\
MLP          && 0.908 & 0.808 & 0.639 && 0.932 & 0.757 & 0.514 \\
\hline
\textbf{Diff. High} && \textcolor{gray}{0.979} & \textcolor{gray}{0.500} & \textcolor{gray}{0.000} && \textcolor{gray}{0.983} & \textcolor{gray}{0.500} & \textcolor{gray}{0.000} \\
Log. Regr.   && 0.979 & 0.684 & 0.000 && 0.983 & 0.789 & 0.000 \\
Rnd. Forest  && 0.979 & 0.706 & 0.000 && 0.983 & 0.618 & 0.000 \\
Grad. Boost. && 0.979 & 0.688 & 0.000 && 0.983 & 0.727 & 0.000 \\
MLP          && 0.979 & 0.681 & 0.021 && 0.983 & 0.675 & 0.000 \\
\hline
\end{tabular}
\label{table_7}
\end{table}